%% file: main.tex
\documentclass[10pt,twocolumn,letterpaper]{article}
\usepackage[pagenumbers]{cvpr} 

\usepackage{duckuments}
\usepackage{cuted}
\usepackage{multirow}
\usepackage{graphicx}
\usepackage{makecell}
\usepackage[table]{xcolor}

\usepackage{pifont}

\definecolor{color3}{RGB}{255, 255, 200}
\definecolor{color2}{RGB}{255, 220, 200}
\definecolor{color1}{RGB}{255, 181, 163}

\newcommand{\sbest}{\cellcolor{color2}}
\newcommand{\tbest}{\cellcolor{color3}}

\definecolor{cvprblue}{rgb}{0.21,0.49,0.74}
\usepackage[pagebackref,breaklinks,colorlinks,allcolors=cvprblue]{hyperref}

\def\paperID{ } 
\def\confName{CVPR}
\def\confYear{2026}

\title{RetimeGS: Continuous-Time Reconstruction of 4D Gaussian Splatting}

\author{%
  Xuezhen Wang$^1$ \quad
  Li Ma$^2$ \quad
  Yulin Shen$^3$ \quad
  Zeyu Wang$^{1,3}$ \quad
  Pedro V. Sander$^1$ \\
  $^1$HKUST \quad
  $^2$Netflix Eyeline Labs \quad
  $^3$HKUST(GZ)
}

\begin{document}
 \twocolumn[{%
 \renewcommand\twocolumn[1][]{#1}%
 \maketitle
 \vspace{-2em}
 \centering
 \includegraphics[width=\linewidth]{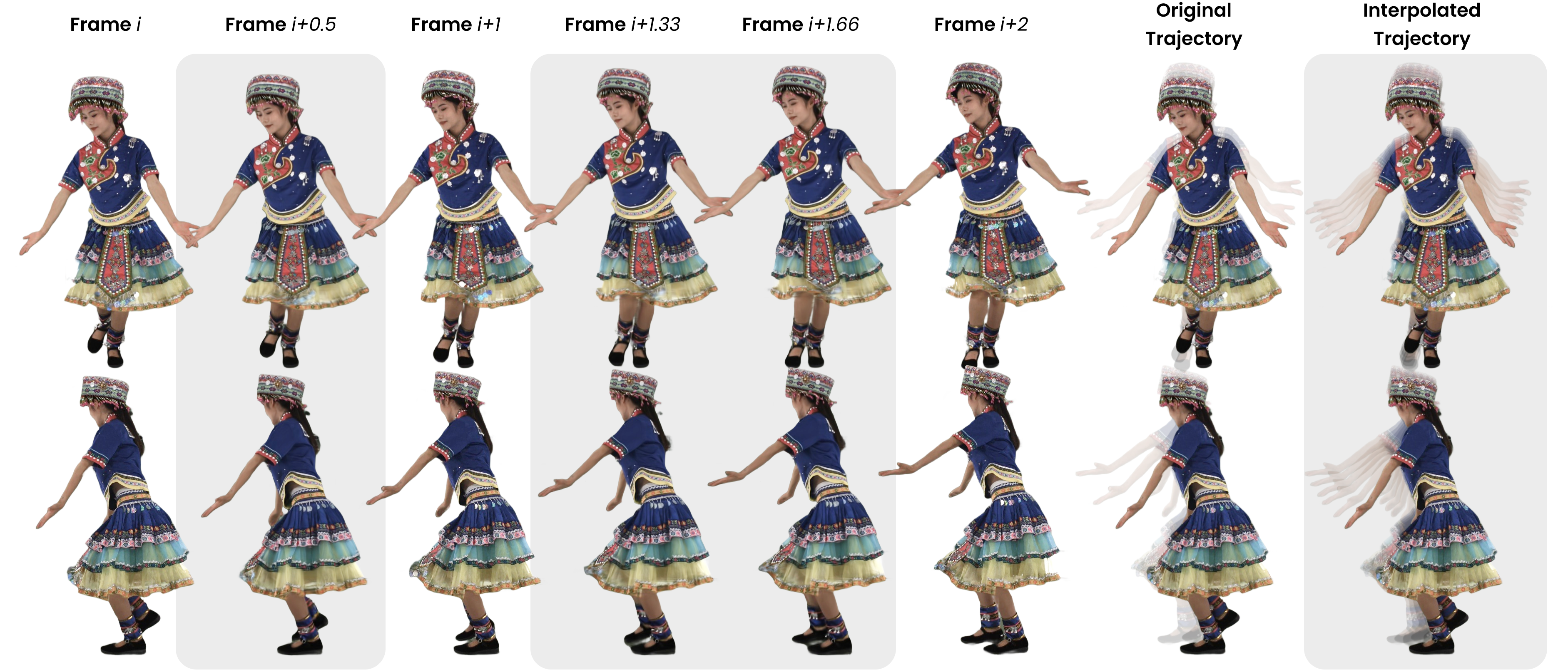}
 \captionof{figure}{
 We introduce a 4DGS representation with tailored training strategies that enables interpolating arbitrary intermediate frames, even under relatively large inter-frame motion.
 Our method reconstructs high-quality frames in challenging scenarios characterized by non-rigid deformations, complex textures, and visibility changes. Project page: \url{https://william-wang2.github.io/RetimeGS/}.\vspace{1em}
 }
 \label{fig:teaser}
 }]

 \input{sec/0_abstract}
 \input{sec/1_introduction}

 \input{sec/2_relatedwork}
 \input{sec/3_methods}

 \input{sec/4_experiment}
 \input{sec/5_conclusion}
 \input{sec/ack}
{
    \small
    \bibliographystyle{ieeenat_fullname}
    \bibliography{main}
}

\input{sec/X_suppl}

\end{document}

%% file: sec/0_abstract.tex
\begin{abstract}
Temporal retiming, the ability to reconstruct and render dynamic scenes at arbitrary timestamps, is crucial for applications such as slow-motion playback, temporal editing, and post-production. However, most existing 4D Gaussian Splatting (4DGS) methods overfit at discrete frame indices but struggle to represent continuous-time frames, leading to ghosting artifacts when interpolating between timestamps. We identify this limitation as a form of temporal aliasing and propose RetimeGS, a simple yet effective 4DGS representation that explicitly defines the temporal behavior of the 3D Gaussian and mitigates temporal aliasing. To achieve smooth and consistent interpolation, we incorporate optical flow-guided initialization and supervision, triple-rendering supervision, and other targeted strategies. Together, these components enable ghost-free, temporally coherent rendering even under large motions. Experiments on datasets featuring fast motion, non-rigid deformation, and severe occlusions demonstrate that RetimeGS achieves superior quality and coherence over state-of-the-art methods.
\vspace{-1em}
\end{abstract}

%% file: sec/1_introduction.tex
 \section{Introduction}
\label{sec:introduction}
High-fidelity dynamic scene reconstruction from multi-view imagery is a fundamental problem in computer vision and computer graphics, with broad applications in virtual reality (VR), film production, and immersive telepresence. 
A common requirement across these applications is precise retime control, which demands rendering a dynamic scene at arbitrary timestamps and producing temporally coherent results far beyond the discrete frame indices of the original capture. Such control enables smooth slow-motion playback, supports the high frame rates required for comfortable VR rendering~\cite{wang2023effect}, and facilitates complex visual effects (VFX) such as speed ramps and bullet time. This typically requires generating continuous intermediate frames between the discrete input frames.

Recently, there has been significant progress in 4D reconstruction using Gaussian Splatting–based representations, owing to their efficiency and high rendering quality. 
Unfortunately, these methods focus primarily on reconstructing a dynamic scene only at discrete input timestamps, and are not optimized for interpolated intermediate times. As a result, when forced to render at floating-point timestamps, they often exhibit various artifacts.
Based on how these methods parameterize the temporal change, they can be broadly grouped into two categories.
A line of research models scene geometry and appearance within a canonical space, leveraging
deformation fields~\cite{bae2024per, guo2024motion, labe2024dgd, liang2025gaufre, lu20243d, shaw2024swings, wu20244d, yang2024deformable, huang2024sc, zhu2024motiongs}, control points~\cite{park2025splinegs}, or physical constraints~\cite{luiten2024dynamic} to capture dynamics.
However, these methods assume that dynamics arise mainly from geometric motion, and therefore struggle when object visibility or textural appearance changes over time. Moreover, they rely on precise correspondence estimation, which becomes unreliable under large motions or limited inter-frame overlap.
As a result, the same primitives may accumulate signals from spatially misaligned regions caused by incorrect correspondence matching, leading to visual artifacts and erroneous trajectories. 

\begin{figure}[t]
\centering
\includegraphics[width=\linewidth]{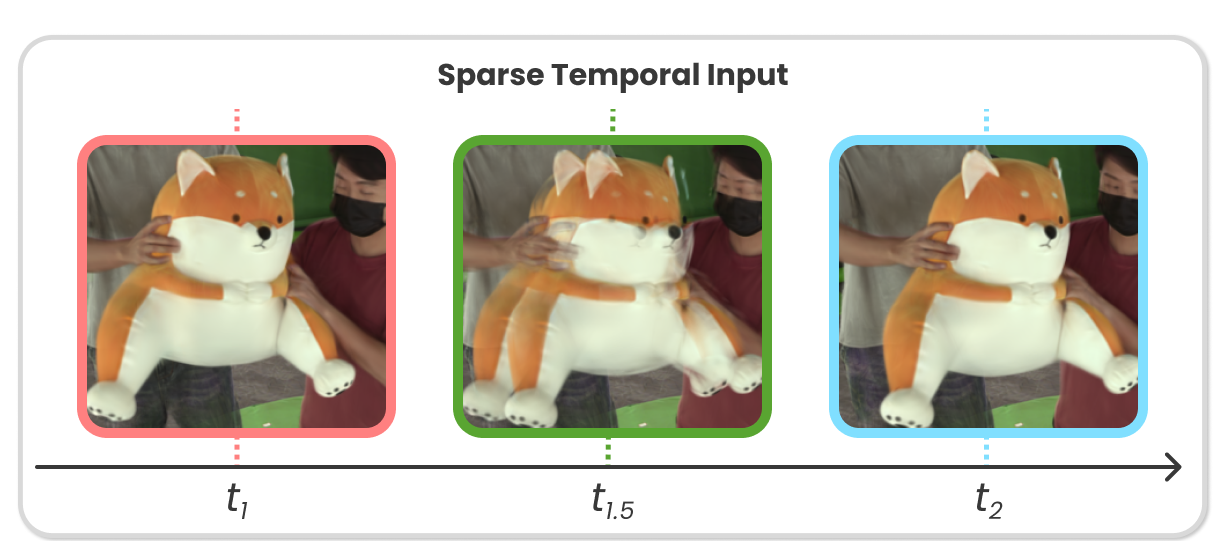}
\vspace{-2em}
\caption{Illustration of temporal overfitting to input frames $t_1$ and $t_2$, causing ghosting at $t_{1.5}$ in 4D primitive-based methods.
}
\vspace{-1em}
\label{fig:illustration_4d}
\end{figure}

Another line of research, which has recently gained widespread adoption, employs 4D primitives to represent dynamic scenes. In these approaches, opacity is typically decomposed into multiple components: a base (or native) opacity, a spatial 3D Gaussian opacity conditioned on time (characterized by scale and rotation that define the covariance), and a temporal opacity modeled using a 1D Gaussian~\cite{wang2025freetimegs, yang2023real, li2024spacetime, xu2024representing, duan:2024:4drotorgs} or other parametric distributions such as constant temporal window with Gaussian fall-off at the boundaries~\cite{lee2024fully}. This formulation allows the scene to be flexibly represented by Gaussians that dynamically appear and disappear according to visibility and appearance changes.
Yet, temporal opacity is freely optimized using supervision only at integer timestamps without any regularization.
As a result, the learned opacity can overfit to discrete frames and become temporally aliased (\eg collapsing to sub-frame temporal support), causing ghosting artifacts when rendering intermediate frames, typically manifested as static semi-transparent overlapping structures from adjacent input frames (see~\Cref{fig:illustration_4d}). This is less problematic for small-motion~\cite{li2022neural} or high-FPS~\cite{wang2025freetimegs} datasets, but can struggle for data with large motion. Analogous to 3D Mip-Splatting~\cite{yu2024mip}, which addresses the problem of spatial aliasing, an intuitive solution for this representation is to apply a low-pass filter to the temporal opacity, effectively widening the primitive. However, this approach introduces a new challenge: the stretched Gaussian distribution requires accurate trajectory estimation across many frames. Failure to do so leads to another form of ghosting artifacts, arising from inconsistent trajectories among different primitives.

The aforementioned limitations motivate the core design principles of our representation. Specifically, it should (i) dynamically appear and disappear to capture variations in appearance and visibility of dynamic content, (ii) be regularized to prevent collapse under sparse temporal sampling, and (iii) maintain accurate and consistent trajectories throughout its duration to avoid ghosting artifacts.
To realize these principles, our representation enforces that a single set of primitives fully explains two adjacent input frames and the interval between them via a short-tailed temporal opacity. We model the trajectory across this interval with a Catmull–Rom spline~\cite{catmull1974class}, whose parameters are supervised by bidirectional optical flow. To avoid redundant representations in static regions, the duration of each primitive is dynamically learned to extend across multiple frames when it can reliably represent them. The entire representation is optimized using triple-rendering supervision, a flow-based initialization strategy and a duration-weighted relocation strategy, enabling smooth interpolation under low frame rates and large motion. Experiments on real-world challenging datasets featuring fast motion, non-rigid deformations, complex textures and visibility changes demonstrate superior performance compared to existing baselines.


%% file: sec/2_relatedwork.tex
\section{Related Work}
\label{sec:ralatedwork}
\subsection{Dynamic Scene Reconstruction}
Since dynamic scene reconstruction has been extensively studied, we refer readers to~\cite{zhu2025dynamic} for a comprehensive survey and focus here on the most relevant methods.

As discussed in~\Cref{sec:introduction}, our design is motivated by two main categories of optimization-based Gaussian Splatting approaches. GaussianFlow~\cite{gao2024gaussianflow} first introduces trajectory supervision via optical flow. However, without explicit regularization on temporal opacity, the primitives still cluster around input frames and rapidly dissolve in intermediate ones, resulting in similar ghosting artifacts. Our representation addresses this issue by regularizing the temporal opacity and parameterizing the primitives such that the primitives can jointly leverage two forward and two backward optical flows to learn a smooth spline trajectory. This design eliminates the linearity bias inherent to optical flow, which is limited to pairwise frame correspondence.
SplineGS~\cite{park2025splinegs} parameterizes trajectories via splines driven by control points corresponding to 2D tracks per frame~\cite{karaev2024cotracker}. SoM~\cite{som2024} and MoSca~\cite{Lei_2025_CVPR}, on the other hand, use sparse motion bases to model trajectories and rely on interpolation for the dense deformation field. Nevertheless, they are tailored to monocular 4D reconstruction. A concurrent work, TrackerSplat~\cite{Yin2025TrackerSplat}, focuses on reconstructing scenes with large motions, yet it is not designed for the interpolation.

Recently, several feed-forward 4D reconstruction methods have been proposed~\cite{ren2024l4gm, xu20254dgt, lin2025dgs, wu2025streamsplat, hu2025forge4d}. However, some of these approaches are constrained by their training datasets—for instance, L4GM~\cite{ren2024l4gm} exhibits poor generalization to real-world subjects, while Forge4D~\cite{hu2025forge4d} is trained exclusively on human data. Others primarily target monocular 4D reconstruction~\cite{xu20254dgt, lin2025dgs, wu2025streamsplat}. Our method leverages all available dense views for better quality and generalizes effectively across diverse datasets and scene types.

\subsection{4D Scene Interpolation}
When only a single view is available, our problem degenerates into video frame interpolation (VFI), a widely studied area. We refer readers to the recent survey~\cite{kye2025acevfi} for a comprehensive overview. A straightforward approach is to generate unseen intermediate frames independently for each video using VFI and then lift them to 3D. However, this is challenging since reconstruction requires both spatial and temporal consistency. 4DSlomo~\cite{chen20254dslomo} takes a step in this direction by trading training views for frame rate and employing a video model~\cite{wan2025} to hallucinate unseen viewpoints, which requires additional setup for asynchronous capture. In-2-4D~\cite{nag20252} tackles 3D start–end frame interpolation with large input disparities, but its reliance on priors such as video models limits generalization to scenes that those models can faithfully represent. All these methods constrain the number of interpolated frames to those produced by the underlying 2D interpolation, whereas our approach can generate arbitrary intermediate frames after reconstruction.
Other methods, such as PAPR~\cite{peng2024papr} and the recent GMC~\cite{GlobalMotionCorresponder}, focus on establishing correspondences for start–end frame interpolation, a different task that becomes prohibitively expensive when applied to an entire sequence.

%% file: sec/3_methods.tex
\section{Methods}
\label{sec:methods}

\begin{figure*}[t]
\centering
\includegraphics[width=\linewidth]{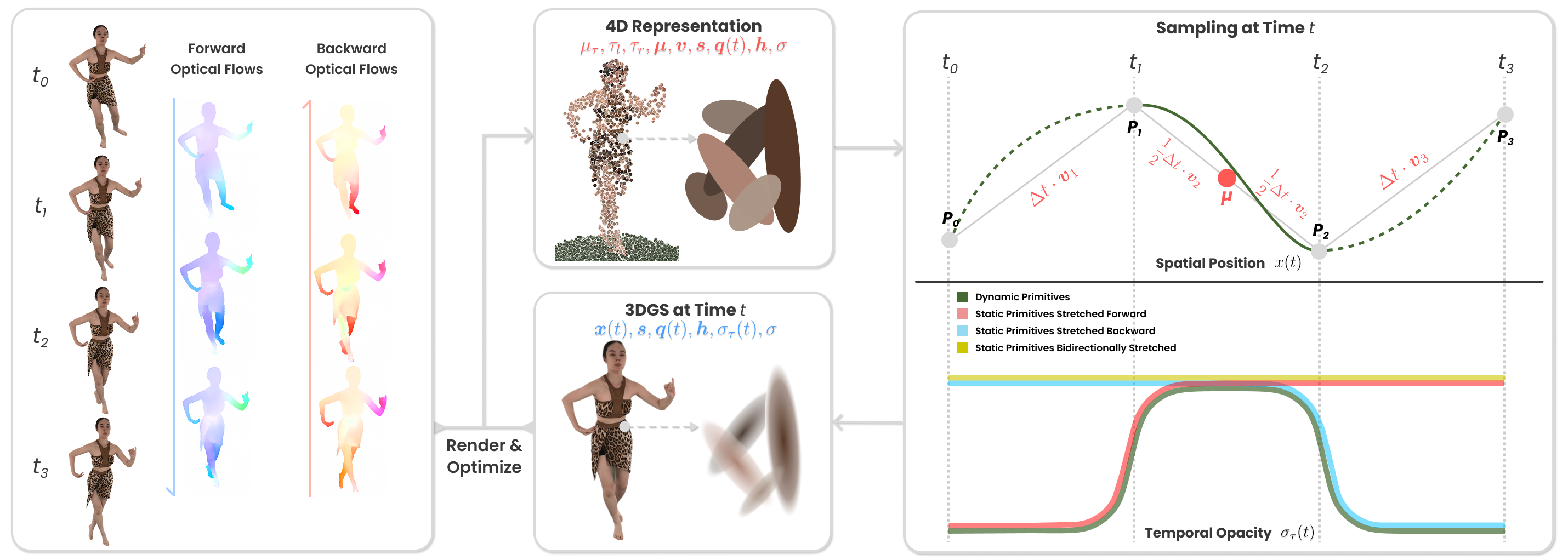}
\vspace{-2em}
\caption{\textbf{Pipeline Overview.} We represent a dynamic scene using a novel 4D representation that combines regularized temporal opacity with smooth spline-based spatial positioning. By leveraging tailored training strategies using RGB images and bidirectional optical flow, our method can reconstruct arbitrary intermediate frames under sparse temporal sampling and large motion.}\label{fig:pipeline}
\end{figure*}

Our method utilizes multi-view videos and corresponding optical flows, derived from off-the-shelf WAFT~\cite{wang2025waftwarpingalonefieldtransforms}, as input to reconstruct a 4D scene at any time, allowing interpolation of arbitrary frames between the input views. Figure~\ref{fig:pipeline} illustrates the overall pipeline of the method.

\subsection{Preliminaries}
For ease of presentation, we formalize these inputs as follows. 
Let $\mathcal{C}=\{c_1,\dots,c_{|C|}\}$ be the set of $C$ camera views and $\mathcal{T}=\{t_1,\dots,t_{|T|}\}$ the set of time steps, where $t_{i+1}-t_i=\Delta t$ for all valid $i$.
A multi-view video is a tensor $\mathbf{V}\in\mathbb{R}^{C\times T\times H\times W\times3}$, where $\mathbf{V}_{c,t}\in\mathbb{R}^{H\times W\times3}$ is the RGB image from camera $c$ at time $t$. 
The corresponding forward multi-view optical flow is $\mathbf{F}^{\mathrm{fwd}}\in\mathbb{R}^{C\times(T-1)\times H\times W\times2}$, where $\mathbf{F}^{\mathrm{fwd}}_{c,t}\in\mathbb{R}^{H\times W\times2}$ gives the per-pixel 2D motion field from $\mathbf{V}_{c,t}$ to $\mathbf{V}_{c,t+1}$ for camera $c$. 
Analogously, the backward multi-view optical flow is $\mathbf{F}^{\mathrm{bwd}}\in\mathbb{R}^{C\times(T-1)\times H\times W\times2}$, where $\mathbf{F}^{\mathrm{bwd}}_{c,t}\in\mathbb{R}^{H\times W\times2}$ gives the per-pixel 2D motion field from $\mathbf{V}_{c,t}$ back to $\mathbf{V}_{c,{t-1}}$.

\subsection{4D Representation Formulation}
\label{sec:representation}

The original 3DGS~\cite{10.1145/3592433} primitives are represented by $(\boldsymbol{x}, \boldsymbol{s}, \boldsymbol{h}, \boldsymbol{q}, \sigma)$. To support dynamic scenes, we extend the parameters of each Gaussian primitive to:
\begin{equation}
\bigl(\mu_{\tau},\ \tau_{l},\ \tau_{r},\ \boldsymbol{\mu},\ \boldsymbol{v},\ \boldsymbol{s},\ \boldsymbol{q}(t),\ \boldsymbol{h},\ \sigma\bigr),
\end{equation}
from which we can get the corresponding 3DGS primitives with $(\boldsymbol{x}(t), \boldsymbol{s}, \boldsymbol{q}(t), \boldsymbol{h}, \sigma_{\tau}(t), \sigma)$ at time $t$.
$\mu_{\tau}$ is the temporal mean, and $\tau_{l}$ and $\tau_{r}$ represent the left and right temporal boundaries, respectively, which are used to define the temporal opacity $\sigma_{\tau}(t)$. 
$\boldsymbol{\mu}$ denotes the pseudo spatial mean, and $\boldsymbol{v} = (\boldsymbol{v}_1, \boldsymbol{v}_2, \boldsymbol{v}_3)$ is the velocity components. $\boldsymbol{\mu}$ and $\boldsymbol{v}$ are combined to specify a spline for the spatial trajectory to get $\boldsymbol{x}(t)$. 
The parameter $\boldsymbol{s}$ represents the anisotropic scale, while $\boldsymbol{q}(t)$ is a quaternion denoting the rotation. 
The coefficients $\boldsymbol{h}$ correspond to the spherical harmonics used for color representation, and $\sigma$ specifies the base opacity.

At any time $t$ for a primitive $p$, following STGS~\cite{li2024spacetime}, the rotation $\boldsymbol{q}_p(t)$ is modeled as a low-order polynomial in time. 
The effective opacity $\sigma_p(\boldsymbol{x}, t)$ contributed by primitive $p$ at spatial location $\boldsymbol{x}$ and time $t$ is
\begin{equation}
    \sigma_{\tau,p}(t)\,\sigma_p\,\exp\!\bigl(-\tfrac{1}{2}(\boldsymbol{x}-\boldsymbol{x}_p(t))^{\!\mathsf{T}}\boldsymbol{\Sigma}_p(t)^{-1}(\boldsymbol{x}-\boldsymbol{x}_p(t))\bigr),
\end{equation}
where $\boldsymbol{\Sigma}_p(t) = \boldsymbol{R}_p(t)\,\mathrm{diag}(\boldsymbol{s}_p^{2})\,\boldsymbol{R}_p(t)^{\mathsf{T}}$is the time-varying covariance obtained by rotating and scaling the base Gaussian of primitive $p$, with $\boldsymbol{R}_p(t)=\mathrm{QuatToMat}\bigl(\boldsymbol{q}_p(t)\bigr)$.
The color of primitive $p$ at time $t$ is evaluated from its spherical harmonics as $\boldsymbol{c}_p(t) = \mathrm{SH}\bigl(\boldsymbol{h}_p,\,\boldsymbol{d}_p(t)\bigr)$, where $\boldsymbol{d}_p(t)$ is the viewing direction from the camera to $\boldsymbol{x}_p(t)$. 
Following the standard Gaussian Splatting formulation~\cite{10.1145/3592433}, the contributions of all primitives are then projected, depth-sorted, and alpha-composited to render the final image at time $t$.

In the following sections, we focus on explaining our design for obtaining the temporal opacity $\sigma_{\tau}(t)$ and the spatial mean $\boldsymbol{x}(t)$ from per-primitive parameters.

\textbf{Temporal Opacity.} 
    For interpolation, as discussed in \Cref{sec:introduction}, our primitives must dynamically appear and disappear to combat the limitations of deformation-based methods, while being regularized to span the duration between input frames. This prevents them from degenerating and clustering around sparse temporal inputs, overcoming the limitations of 4D primitive-based approaches. Moreover, the duration of each primitive must align with the explicitly supervised trajectory, implying that it cannot rely on a stretched temporal distribution with strong support across many frames (\eg, a Gaussian), since accurate trajectory estimation across multiple frames inherently shares the same correspondence challenges as deformation-based methods.

Specifically, at initialization, the temporal mean $\mu_{\tau}$, offsets $\tau_{l}$ and $\tau_{r}$ are non-optimizable and defined as:
\begin{equation}
    \mu_{\tau} = \frac{t_i + t_{i+1}}{2},\quad \tau_{l}=\tau_{r}=\frac{\Delta t}{2}, \quad t_i, t_{i+1} \in \mathcal{T}.
\end{equation}
The temporal opacity $\sigma_{\tau}(t)$ is formulated as the product of two sigmoid functions centered at the left and right temporal boundaries defined by the temporal mean $\mu_{\tau}$ and offsets $\tau_{l}$ and $\tau_{r}$, respectively.
This design allows each primitive to smoothly fade in and out within its temporal range with a short-tailed kernel.
At boundaries, however, there is no primitive beyond the video duration to blend with.
To prevent an undesired drop in visibility near these global limits, the corresponding sigmoid function is replaced by a constant value of $1$, ensuring that the primitive remains fully visible up to the boundary.
Formally, we define
\begin{equation}
\sigma_{\tau}(t) =
\tilde{\psi}_{l}\!\left(\frac{t - (\mu_{\tau} - \tau_{l})}{\gamma}\right)
\tilde{\psi}_{r}\!\left(\frac{(\mu_{\tau} + \tau_{r}) - t}{\gamma}\right),
\end{equation}
where $\tilde{\psi}_{l}$ and $\tilde{\psi}_{r}$ are boundary-aware sigmoid functions defined compactly as
\begin{equation}
\tilde{\psi}_{s}(x) =
\begin{cases}
1, & \text{if } 
\begin{cases}
\mu_\tau-\tau_{l} < \epsilon, & s = l,\\
\mu_\tau+\tau_{r} > t_T -\epsilon, & s = r,
\end{cases}\\[6pt]
\psi(x), & \text{otherwise.}
\end{cases}
\end{equation}
Here, $\psi(\cdot)$ denotes the sigmoid function, and $\gamma$ is a hyperparameter controlling the smoothness of temporal transitions.
Intuitively, this design keeps each set of primitives centered and active between two input frames, and supervises it using both frames, allowing smooth rendering of intermediate frames. Around each input frame, two neighboring sets of primitives blend in and out, ensuring seamless transitions.

Furthermore, to leverage the same advantages offered by deformation-based methods, which require fewer primitives, as will be explained later in~\Cref{sec:stretch_and_relocate}, $\tau_{l}$ and $\tau_{r}$ are periodically extended to $(\frac{1}{2}+k)\Delta t$ for some positive integer $k$ during training when certain criteria are met. This allows static primitives to persist for longer durations, and once these static parts begin to move, which are automatically determined by the stretched boundary, they rapidly dissolve as a new set of primitives gradually fades in. \Cref{fig:pipeline} shows dynamic primitives as well as primitives that are periodically detected as static and temporally stretched, with a temporal mean centered at $\frac{t_1 + t_2}{2}$.

\textbf{Spatial Mean.}
Regularizing temporal opacity alone is insufficient for accurate interpolation. Without additional supervision signals, sparse temporal inputs yield minimal or no content overlap between adjacent frames for moving objects, which prevents RGB supervision from learning reliable correspondences. In the absence of accurate trajectories, the primitives receive inconsistent signals from the two input frames, resulting in suboptimal representations. Furthermore, under sparse temporal input, assuming linear velocity for large motions results in piecewise-linear motion artifacts. These motivate our decision to model the spatial mean $\boldsymbol{x}(t)$ using a Catmull-Rom spline~\cite{catmull1974class} and explicitly supervise its parameters by a bidirectional optical flow.

As shown in~\Cref{fig:pipeline}, intuitively, for a primitive $p$ with a temporal mean $\mu_{\tau,p} = (t_i + t_{i+1})/2$, the velocity $\boldsymbol{v}_{2,p}$ represents the linear velocity between frames $t_i$ and $t_{i+1}$ derived from 3D correspondence. Similarly, $\boldsymbol{v}_{1,p}$ is the linear velocity from $t_{i-1}$ to $t_i$, and $\boldsymbol{v}_{3,p}$ is the linear velocity from $t_{i+1}$ to $t_{i+2}$. The pseudo-mean $\boldsymbol{\mu}_p$ represents the position at $\mu_{\tau,p}$, assuming linear motion from $t_i$ to $t_{i+1}$ between the correspondences. If $\boldsymbol{v}_{1,p}$ or $\boldsymbol{v}_{3,p}$ is unavailable at temporal boundaries, we fall back to $\boldsymbol{v}_{2,p}$.

The inner control points, which the spline interpolates exactly, correspond to the positions at $t_i$ and $t_{i+1}$.
\begin{equation}
\label{eq:middlepoints}
\boldsymbol{p}_{1,p} = \boldsymbol{\mu}_p - \frac{1}{2}\Delta t \cdot \boldsymbol{v}_{2,p}, \qquad \boldsymbol{p}_{2,p} = \boldsymbol{\mu}_p + \frac{1}{2}\Delta t \cdot \boldsymbol{v}_{2,p}
\end{equation}
The outer control points determine the curvature at the inner points using velocities from adjacent intervals.
\begin{equation}
\label{eq:outerpoints}
\boldsymbol{p}_{0,p} = \boldsymbol{p}_{1,p} - \Delta t \cdot \boldsymbol{v}_{1,p}, \qquad 
\boldsymbol{p}_{3,p} = \boldsymbol{p}_{2,p} + \Delta t \cdot \boldsymbol{v}_{3,p}
\end{equation}
Utilizing these four control points, the Catmull-Rom spline~\cite{catmull1974class} smoothly interpolates the trajectory $\boldsymbol{x}_p(t)$ for $t \in [t_i, t_{i+1}]$, which is the active duration for primitive $p$ if it is dynamic. Notably, for static primitives, their velocity vectors $\boldsymbol{v}_p$ are approximately zero. Consequently, even if their temporal support is stretched, the trajectory remains valid as extrapolation beyond $t_i$ and $t_{i+1}$ yields a consistent static position. The training on these parameters will be explained in~\Cref{sec:bidirectionalflow}. In experiments, we observe that optimizing the pseudo mean and velocity components as parameters is much easier than optimizing the four control points, even though they are mathematically equivalent.

\subsection{Training Strategies}
\label{subsec:training}
In this subsection, we introduce four complementary training strategies that jointly optimize our 4D representation to enable rendering and interpolation at arbitrary frames.
\subsubsection{Bidirectional Flow Trajectory Supervision}\label{sec:bidirectionalflow}

We leverage optical flow to establish coarse correspondences between frames, which are used to supervise the trajectory-related parameters of each primitive: its pseudo-mean $\boldsymbol{\mu}$ and velocity components $\boldsymbol{v} = (\boldsymbol{v}_1, \boldsymbol{v}_2, \boldsymbol{v}_3)$. These parameters are optimized through backpropagated gradients from the four control points defined in \Cref{eq:middlepoints} and \Cref{eq:outerpoints}. Consider supervision at frame $t_i$, for primitives whose temporal mean satisfies $\mu_{\tau} = \frac{t_{i-1} + t_i}{2}$, \ie, the group temporally preceding $t_i$ (if any), we compute the projected 3D flow from $\boldsymbol{p}_2$ to $\boldsymbol{p}_1$ and from $\boldsymbol{p}_2$ to $\boldsymbol{p}_3$ under camera view $c$. These are treated as two-dimensional feature vectors to rasterize backward and forward flow maps, which are supervised against the ground-truth optical flows $\mathbf{F}^{\mathrm{bwd}}_{c,t_i}$ and $\mathbf{F}^{\mathrm{fwd}}_{c,t_i}$ via per-pixel losses. Similarly, for primitives whose temporal mean satisfies $\mu_{\tau} = \frac{t_i + t_{i+1}}{2}$, \ie, the group temporally following $t_i$ (if any), we project 3D flow from $\boldsymbol{p}_1$ to $\boldsymbol{p}_2$ and from $\boldsymbol{p}_1$ to $\boldsymbol{p}_0$, rasterize the flow maps, and supervise them against $\mathbf{F}^{\mathrm{fwd}}_{c,t_i}$ and $\mathbf{F}^{\mathrm{bwd}}_{c,t_i}$, respectively.

When rendering the flow maps, we divide the temporal opacity by the corresponding $\sigma_\tau(t_i)$ because two sets of active primitives are rendered separately at $t_i$. This normalization ensures that the combined contribution of dynamic primitives remains approximately consistent with the original temporal composition at $t_i$ and also recovers the correct temporal opacity during the interval between input frames, where the temporal opacity is approximately $1$ for both dynamic and static primitives that are active.

Once the trajectory parameters stabilize during training, we gradually reduce the flow supervision learning rate toward zero, relying increasingly on RGB supervision for fine-grained refinement of primitives.

\subsubsection{Triple Rendering}
\label{sec:triple-rendering}
Ideally, at the two adjacent intervals around any input frame $t_i \in \mathcal{T}$, \ie, $[t_{i-1}, t_i]$ and $[t_{i}, t_{i+1}]$ (if any), primitives active in that interval (including static primitives which span more than one interval and the dynamic primitives active only in that interval) should be able to explain frame $t_i$ alone faithfully. 
However, in practice, we find that rendering all primitives together reproduces the image at $t_i$, whereas each subset typically reconstructs different regions with uneven coverage. As a result, rendering the subsets individually leads to under-reconstruction in intermediate frames (see~\Cref{fig:ablation_triple_render}).
We propose a simple yet effective approach to address this issue.  
For each interior frame $t_i \in \{2,\ldots,T-1\}$, we render three images: one using all primitives, and two using each set individually with the same temporal opacity compensation as before.  
For the boundary frames $t_1$ and $t_T$, where only one set of primitives exists, we render a single image without opacity compensation.  
All rendered images are supervised against the ground-truth image at $t_i$.

\subsubsection{Dynamic Stretching and Periodic Relocation}
\label{sec:stretch_and_relocate}
As discussed in~\Cref{sec:representation}, to prevent redundant representation of static objects with multiple primitive sets, we periodically stretch the left and right temporal boundaries $\tau_{l}$ and $\tau_{r}$. Precisely, once training stabilizes, primitives with $\mu_\tau = \frac{t_i + t_{i+1}}{2}$ search for their nearest neighbors in adjacent intervals, \ie, those with $\mu_\tau = \frac{t_{i-1} + t_i}{2}$ and $\mu_\tau = \frac{t_{i+1} + t_{i+2}}{2}$ (if any). If the matched primitives have similar degree-0 Spherical Harmonics (base color) and near-zero velocity, their $\tau_{l}$ and $\tau_{r}$ are stretched so that both span the union of their original temporal ranges. After stretching, a primitive is pruned with probability $1 - \frac{1}{k+1}$, where $k$ is the number of times other primitives match it. This encourages removing redundant primitives whose content is now likely to be covered by stretched primitives. Through progressive stretching, the temporal extent of static primitives is extended until motion occurs, at which point a new set of primitives fades in to represent the dynamic region.

We adopt the MCMC strategy~\cite{kheradmand20243d, wang2025freetimegs} to our representation. Primitives with base opacity $\sigma$ below a predefined threshold are periodically moved to regions whose primitives exhibit higher sampling scores, which is defined as
\begin{equation}
    s = \frac{\sigma}{\tau_{l} + \tau_{r}}.
\end{equation}
Intuitively, it reweights the base opacity based on the temporal duration, encouraging relocation to dynamic regions.

\subsubsection{Flow-Aware Initialization}

To demonstrate the robustness of our method and achieve efficient initialization, we use VGGT~\cite{wang2025vggt} without bundle adjustment to estimate unrefined point clouds for each frame $t_i \in \mathcal{T}$ and align the point clouds from VGGT’s world coordinate to the ground-truth camera coordinate system. To obtain a reasonable initial estimate of the parameters for primitives temporally located midway between two input frames, we proceed as follows. For each camera $c_i \in \mathcal{C}$ and frame $t_i$, the points are projected onto the image plane, where the 2D forward flows $\mathbf{F}^{\mathrm{fwd}}_{c_i,t_i}$ and backward flows $\mathbf{F}^{\mathrm{bwd}}_{c_i,t_i}$ are bilinearly interpolated. The 2D flows from all views are then back-projected to 3D and averaged to estimate the forward and backward 3D flows. The average of them is used to initialize all velocity components $\boldsymbol{v} = (\boldsymbol{v}_1, \boldsymbol{v}_2, \boldsymbol{v}_3)$. The pseudo mean $\boldsymbol{\mu}$ for primitives between $t_i$ and $t_{i+1}$ is approximated by displacing the points at $t_i$ and $t_{i+1}$ using the estimated velocity.

%% file: sec/4_experiment.tex
\section{Experiments}
\subsection{Experiment Settings}
Our implementation is based on PyTorch~\cite{NEURIPS2019_9015}, and all experiments are conducted on a single RTX 4090D GPU. Additional details are provided in the supplementary materials.

\textbf{Experimental Setup.} Our dataset comprises 10 scenes from the DNA-Rendering dataset~\cite{2023dnarendering} and 9 scenes we captured in-stage (denoted as the Stage-Capture Dataset). Each 17-frame clip features challenging conditions, including changes in visibility, fast motion, intricate textures, highly non-rigid deformation, and complex interactions. For quantitative evaluation, we train using all cameras and measure performance against the held-out intermediate frame, using three key metrics: PSNR (pixel-level error); SSIM~\cite{wang2004image} (perceptual similarity based on luminance, contrast, and structure); and LPIPS~\cite{zhang2018unreasonable} (deep features aligning with human perceptual judgment). Since stage data contains large static background regions that can disproportionately inflate all metric scores for evaluating interpolated frames, we compute PSNR and SSIM only within the foreground region and mask out the background when calculating LPIPS.

The DNA-Rendering dataset captures dynamic subjects at 15 FPS using 60 4K/2K cameras. We use all its frames for training and qualitative study. We also captured 9 new scenes with 32 synchronous 4K RGB cameras at 22 FPS. From these, we hold out every other frame, effectively using 11 FPS video clips for training. These held-out frames serve as the ground truth for testing temporally interpolated novel views, allowing us to quantitatively and qualitatively evaluate our method and compare against baselines. All data is scaled to 1K resolution for fast training. These datasets contain challenging scenes, such as dancing in a dress with intricate textures, taking off clothes where hand visibility changes, waving sleeves that involve highly non-rigid motion, and kicking a rotating football, which demonstrates complex interactions.

\subsection{Comparison}
\label{sec:comparison}

\begin{figure*}[t]
\centering
\includegraphics[width=\linewidth]{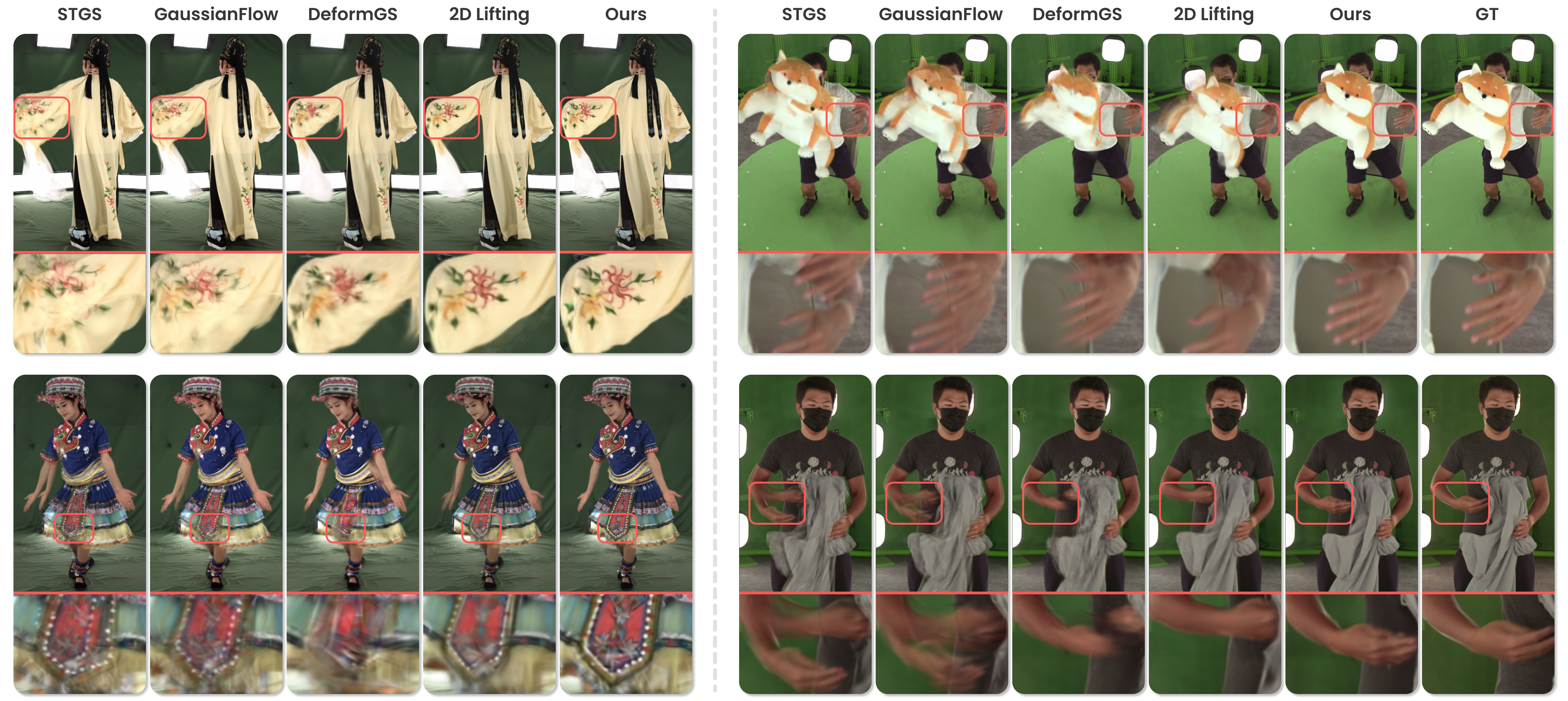}
\vspace{-2em}
\caption{\textbf{Qualitative comparison.} 
Results on DNA-Rendering Dataset~\cite{2023dnarendering} (w/o GT held-out views, left) 
and Stage-Capture Dataset (w/ GT held-out views, right). 
The red boxes show the corresponding zoomed-in views for detailed comparison.}
\label{fig:comparison}
\end{figure*}
\input{table/comparison_result}
We quantitatively compare our method against representative deformation-based~\cite{wu20244d} and 4D primitives-based~\cite{li2024spacetime} methods. We also include GaussianFlow~\cite{gao2024gaussianflow}, which adds optical flow supervision, and a baseline that lifts 2D video interpolation by applying FILM~\cite{reda2022film} independently to each view to reconstruct the middle frames using~\cite{li2024spacetime}. The results in~\Cref{tab:comparison_result} clearly demonstrate that our method outperforms all baselines across all three evaluation metrics.

The advantage of our method is more clearly illustrated through qualitative comparisons. As shown in~\Cref{fig:comparison}, we visualize the interpolated temporal novel views for five different methods. Under large inter-frame displacements, STGS~\cite{li2024spacetime} often produces noticeable ghosting artifacts, as discussed in~\Cref{sec:introduction}. Specifically, it tends to overfit the preceding and subsequent input frames, resulting in overlapping semi-transparent regions for fast-moving parts, such as the yellow sleeves, hands, and dolls

GaussianFlow~\cite{gao2024gaussianflow}, which introduces forward optical flow supervision, alleviates this issue only marginally. Even when trajectories satisfy flow constraints, moving primitives toward other input frames interferes with those already present, and RGB loss consequently drives $\sigma_{\tau}$ to collapse toward a single frame. Intuitively, the optimizer can assign velocities that satisfy flow supervision while still shortening the primitives’ temporal support, since allowing them to persist across frames increases the difficulty of matching RGB signals. Therefore, compared with STGS~\cite{li2024spacetime}, the distance between overlapping objects is reduced, yet ghosting remains visible in the three aforementioned regions.

The deformation-based method~\cite{wu20244d} enforces a single set of primitives to represent all frames in a dynamic scene, which allows it to capture a roughly correct global trajectory compared with STGS~\cite{li2024spacetime}. Nevertheless, in regions with fast motion, complex textures, or visibility changes, establishing detailed correspondences becomes challenging. As shown in~\Cref{fig:comparison}, this method exhibits significant blur and distorted textures. In the bottom-right scene, where the fingers dynamically emerge from the clothing, the method fails to estimate even coarse correspondences, causing the hand to split into two separate parts.

Independently lifting interpolated frames from each view into 3D leads to both view and temporal inconsistencies. On one hand, it can lead to ghosting or blurring artifacts, as evidenced by the faint yellow silhouette near the bottom of the sleeve and the blurred texture on the dress in the left two scenes in~\Cref{fig:comparison}. On the other hand, such inconsistencies may also cause objects to be reconstructed at incorrect spatial positions, \eg, in the right two scenes, the hands appear too close to one of the adjacent input frames.

Our method enforces a single set of primitives to jointly explain at least two frames while leveraging explicit trajectory supervision, making it robust to challenging dynamic scenarios. For instance, in the last scene where fingers dynamically appear, the method uses both flow and RGB information from the next frame to infer the position and appearance of fingers in the intermediate novel frame. Therefore, it achieves faithful finger reconstruction even when the fingers are visible in only one of the adjacent frames.


\subsection{Ablation Study}
\label{sec:ablation}

\begin{figure*}[t]
\centering
\begin{subfigure}[t]{0.35\textwidth}
    \centering
    \includegraphics[height=6cm]{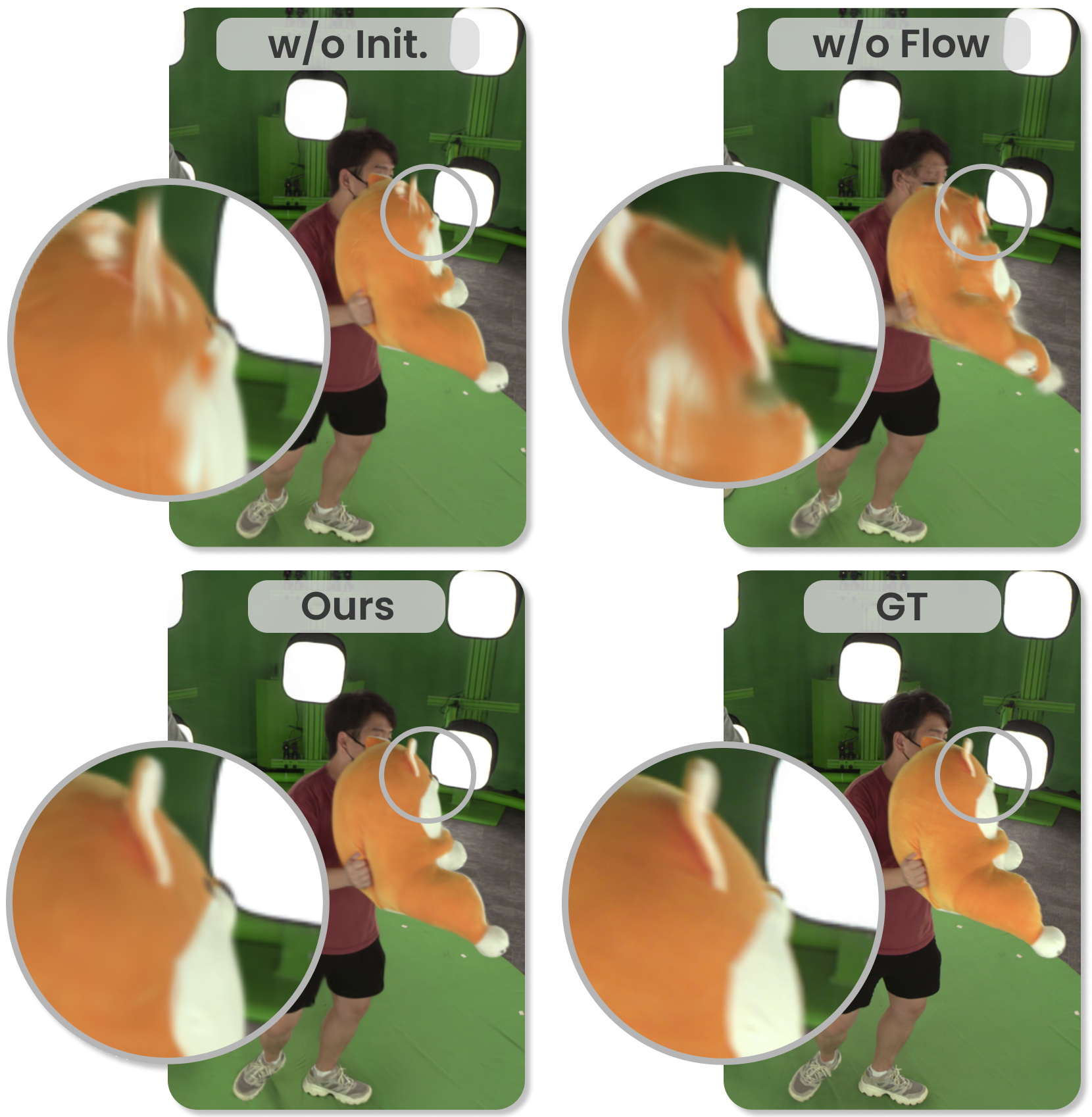}
    \caption{\textbf{Ablation on flow initialization and supervision.}}
    \label{fig:ablation_flow}
\end{subfigure}%
\hspace{22.5pt}%
\begin{subfigure}[t]{0.55\textwidth}
    \centering
    \includegraphics[height=6cm]{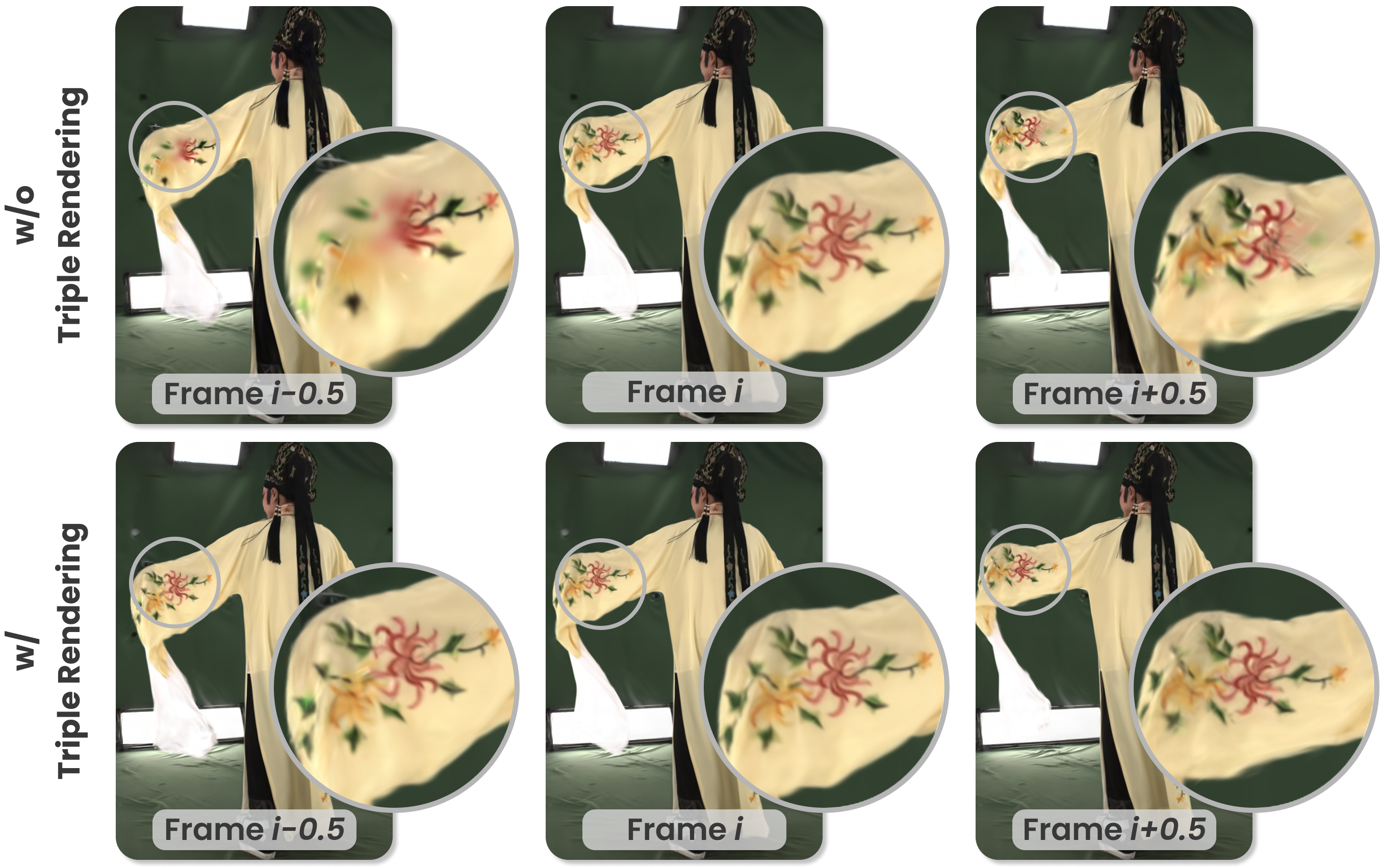}
    \caption{\textbf{Ablation on triple rendering.}}
    \label{fig:ablation_triple_render}
\end{subfigure}
\vspace{-1em}
\caption{ (\ref{fig:ablation_flow})~The texture is distorted when flow-related components are removed. (\ref{fig:ablation_triple_render})~Without triple rendering, the two adjacent primitive groups each capture only part of the content in the input frames they jointly cover: the previous group reconstructs the right texture in the circled region, while the next group reconstructs the left texture.}
\label{fig:ablation_combined}
\vspace{-1em}
\end{figure*}

We conduct quantitative ablation studies across all scenes in our Stage-Capture Dataset using temporal novel views, as summarized in~\Cref{table:ablation}. The results highlight the necessity of each component in our framework. In addition, we provide a qualitative analysis based on interpolated temporal novel views to further illustrate the role and importance of these components.\input{table/ablation_result}

\textbf{Flow Supervision and Flow Initialization.}
We utilize both forward and backward optical flows to establish coarse correspondences for our representation. As shown in~\Cref{fig:ablation_flow}, textures become noticeably distorted for fast-moving objects when either flow supervision or flow-aware initialization is omitted.

\textbf{Triple Rendering.}
As discussed in~\Cref{sec:triple-rendering}, directly rendering all primitives in the input frames and supervising them with ground-truth RGB images, as in previous works, leads to artifacts. Primitives from neighboring temporal intervals jointly reconstruct the input frame but contribute unevenly to spatial regions, resulting in inconsistent reconstruction.
As shown in~\Cref{fig:ablation_triple_render}, dynamic primitives between $t{-}1$ and $t$ miss the left sleeve texture, while those between $t$ and $t{+}1$ miss the right, even though their combined rendering matches the ground truth at $t$.
Our triple-rendering strategy avoids this by requiring each primitive set to explain its corresponding input frames independently.

\begin{figure}[t]
\centering
\includegraphics[width=\linewidth]{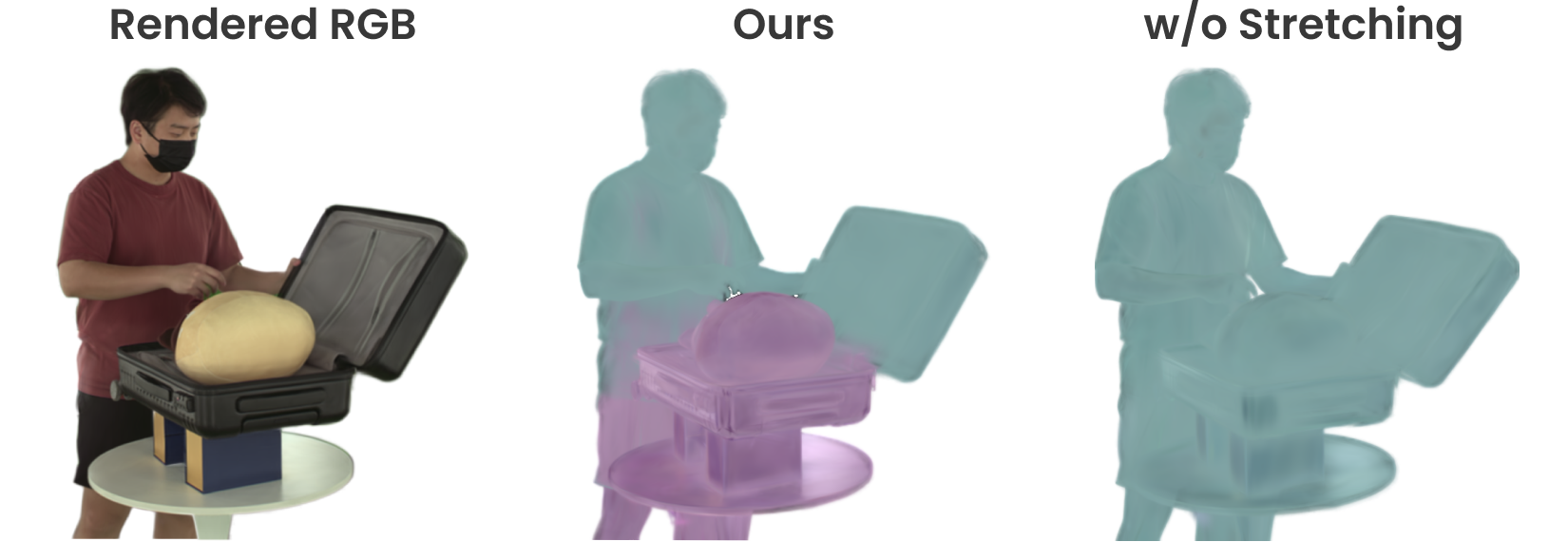}
\vspace{-1.5em}
\caption{
\textbf{Ablation on dynamic stretching.} The magenta is rendered using static stretched primitives, and the teal is rendered using dynamic primitives (with static background removed).}
\label{fig:ablation_stretching}
\vspace{-1em}
\end{figure}

\textbf{Dynamic Stretching.}
We illustrate the effectiveness of our dynamic stretching in~\Cref{fig:ablation_stretching} by rendering primitives that span more than two frames in magenta and those that span exactly two frames in teal (background removed for clarity on static foreground objects). As shown, static regions such as the table, the doll, the lower half of the luggage, and the legs appear in magenta, while the upper body and the top half of the luggage, corresponding to the moving regions, are rendered in teal. Because our MCMC-based density control~\cite{kheradmand20243d,wang2025freetimegs} constrains the total number of Gaussians, dynamic stretching causes the periodic relocation strategy to allocate more primitives to harder-to-reconstruct moving regions, which improves reconstruction quality in dynamic areas. An analysis of the resulting reduction in the effective number of primitives is provided in the supplementary material. 

\begin{figure}[t]
\centering
\includegraphics[width=\linewidth]{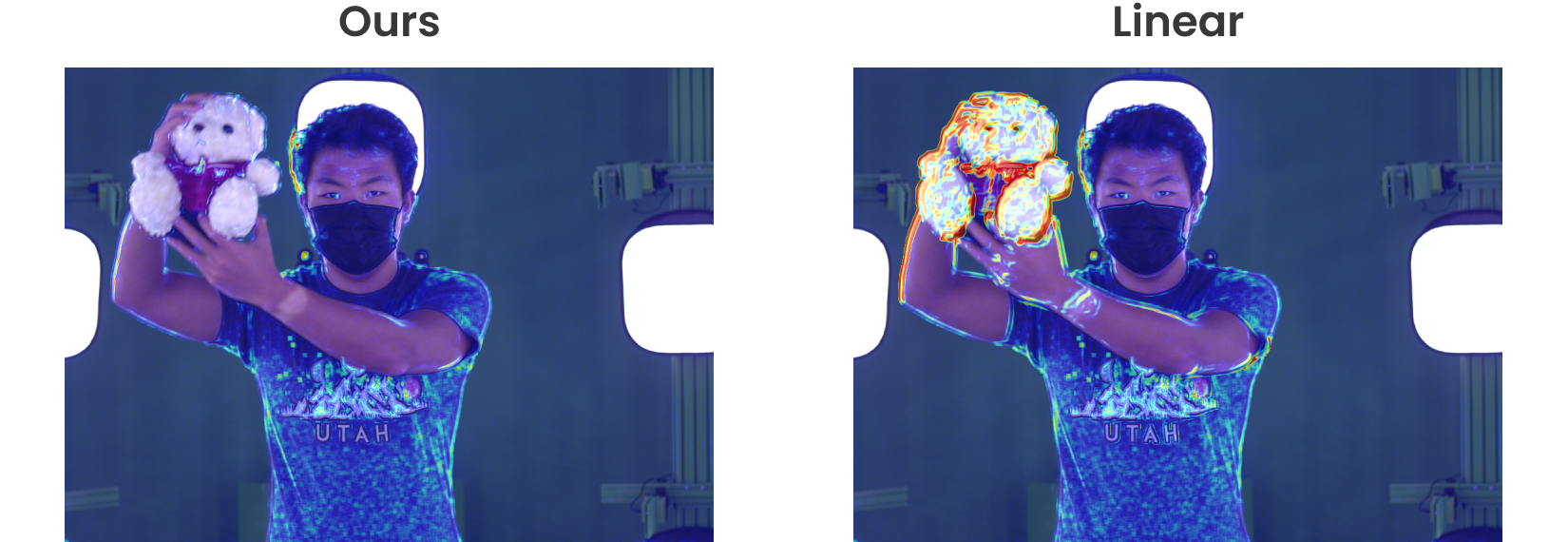}
\vspace{-1.5em}
\caption{
\textbf{Ablation on spline trajectory.} Without spline trajectories, the error heatmap highlights larger errors along the bear’s edges due to imprecise piecewise linear motion.
\vspace{-1em}
}
\label{fig:ablation_trajectory}
\end{figure}

\textbf{Spline Trajectory.} Switching from linear to a spline trajectory produces smoother primitive motion between input frames and avoids abrupt velocity changes when transitioning between dynamic primitive segments. As shown in~\Cref{fig:ablation_trajectory}, the difference heatmap between the ground-truth and interpolated frame highlights pronounced edge errors on the circular moving part under the linear trajectory, which are substantially reduced using the spline trajectory.

%% file: table/comparison_result.tex
\begin{table}[t]
\caption{
\textbf{Quantitative comparisons on the Stage-Capture Dataset, focusing on the foreground region.} Red and yellow cell colors indicate the best and second-best results, respectively.
}
\vspace{-0.5em}
\centering
\begin{tabular}{l ccc}
\toprule
Method & PSNR $\uparrow$ & SSIM $\uparrow$ & LPIPS $\downarrow$ \\
\midrule
Deform-GS~\cite{wu20244d} & 28.45 & 0.867 & 0.0272 \\
STGS~\cite{li2024spacetime} & 25.34 & 0.825 &  0.0357 \\
GaussianFlow~\cite{gao2024gaussianflow} & 25.91 & 0.825 &  0.0339 \\
2D Lifting~\cite{reda2022film, li2024spacetime} & \tbest 28.79 & \tbest 0.886 & \tbest 0.0267 \\
\midrule
Ours & \sbest 30.08 & \sbest 0.904 & \sbest 0.0225 \\
\bottomrule
\end{tabular}
\vspace{-1em}
\label{tab:comparison_result}
\end{table}

%% file: table/ablation_result.tex
\begin{table}[t]
\caption{
\textbf{Quantitative ablation on the Stage-Capture Dataset, focusing on the foreground region.} Red and yellow cell colors indicate the best and second-best results, respectively.
}
\vspace{-0.5em}
\centering
\begin{tabular}{l ccc}
\toprule
Method & PSNR $\uparrow$ & SSIM $\uparrow$ & LPIPS $\downarrow$ \\
\midrule
w/o flow initialization & \tbest 29.69 & \tbest 0.899 & \tbest 0.0227 \\
w/o flow supervision & 27.24 & 0.861 &  0.0282 \\
w/o triple-rendering & 27.16 & 0.849 &  0.0319 \\
w/o dynamic stretching & 28.81 & 0.886 & 0.0247 \\
linear trajectory & 28.50 & 0.884 & 0.0243 \\
\midrule
Ours & \sbest 30.08 & \sbest 0.904 & \sbest 0.0225 \\
\bottomrule
\end{tabular}
\vspace{-1em}
\label{table:ablation}
\end{table}

%% file: sec/5_conclusion.tex
\section{Conclusion}
\label{sec:conclusion}
We introduce RetimeGS, a novel 4DGS representation with tailored training strategies that enable continuous-time reconstruction and rendering at arbitrary timestamps. We identify the drawbacks of the two dominant 4D representation paradigms and propose a new temporal opacity formulation and a spatial mean design accordingly, which address their limitations and integrate the strengths of both approaches. In addition, we propose several essential training strategies, including bidirectional flow trajectory supervision, triple rendering, dynamic stretching, periodic relocation, and flow-aware initialization, all of which are critical for optimizing our representation. Experiments on low-frame-rate datasets with challenging scenarios demonstrate that RetimeGS achieves high-quality continuous interpolation between input frames.

Our method exhibits limitations when applied to videos captured at extremely low frame rates due to the inherent constraints of optical flow. In such cases, significant inter-frame discrepancies reduce the task to a 4D variant of multi-start-end frame interpolation. Addressing this challenge remains an area for future work. We provide further discussion of limitations and future directions in the appendix.


%% file: sec/ack.tex
\section*{Acknowledgement}
\label{sec:acknowledgement}
This research was partially supported by the Hong Kong Research Grants Council under GRF grant (No.~16218824) and by the Guangdong Basic and Applied Basic Research Foundation (No.~2026A1515011138). We are grateful to the Dynamic Reconstruction and Applied Meta Studio (DREAMS) at HKUST(GZ) for providing the stage for dataset collection. Special thanks go to Duotun Wang and Yaodong Yang for their help during data capture. We also thank Yunfan Zeng and Yapeng Meng for insightful discussions throughout the project. Finally, we sincerely thank the anonymous reviewers for their detailed and constructive feedback.


%% file: sec/X_suppl.tex
\clearpage
\setcounter{page}{1}
\maketitlesupplementary

\setcounter{section}{0}      

\renewcommand{\thesection}{\Alph{section}}

\section{Hyperparameter Settings}
The implementation adopts the hyperparameter configurations recommended by gsplat library~\cite{ye2025gsplat}. Opacity regularization and scale regularization are enabled to encourage primitives to maintain low opacity and compact scale, with corresponding weights set to $0.01$ and $0.1$, respectively. The temporal opacity hyperparameter $\gamma$ is set to $0.005$ to promote a short-tail distribution. The MCMC strategy~\cite{kheradmand20243d} relocates primitives every 100 iterations using a minimum opacity threshold of $0.01$. The flow learning rate is initialized at $0.5$ and decays exponentially toward $1 \times 10^{-6}$ after $12{,}000$ iterations. For all Gaussian properties, we apply an exponential decay to the learning rates after $18{,}000$ iterations to encourage stable convergence at the end of training. Additionally, dynamic stretching is applied every $3{,}000$ iterations to adjust the temporal durations of the primitives. All scenes are trained for a total of $20{,}000$ iterations.

\section{More Discussion of Limitations}
\begin{figure}[t]
\centering
\includegraphics[width=\linewidth]{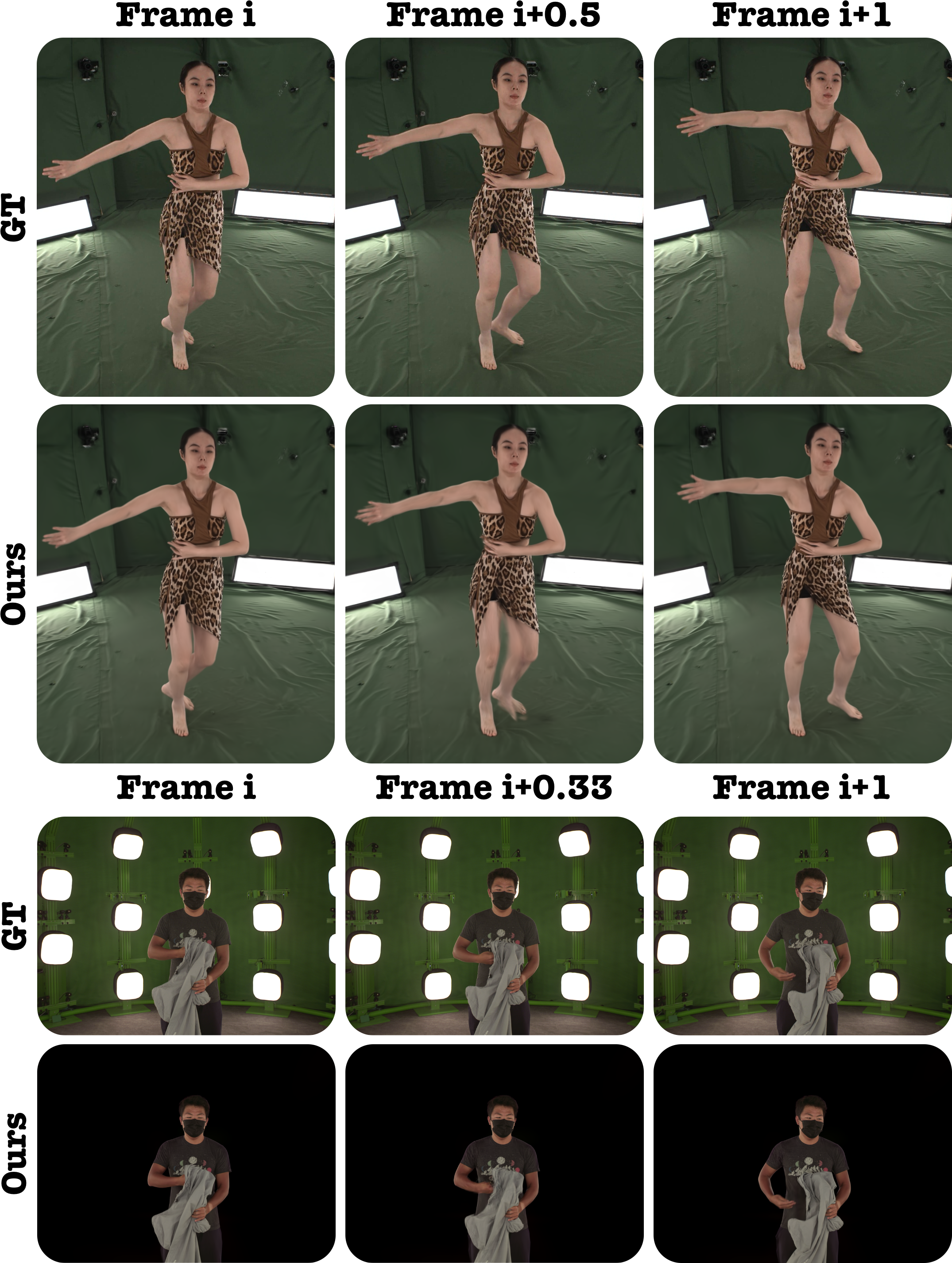}
\vspace{-1.5em}
\caption{
\textbf{Failure case under extremely low capture FPS.} 
Our method struggles to interpolate intermediate frames when the inter-frame motion becomes too large due to low temporal sampling or large motion.
}
\label{fig:fail}
\end{figure}

As noted in~\Cref{sec:conclusion}, our method fails when the inter-frame motion is excessively large or when videos are captured at extremely low frame rates, causing off-the-shelf optical flow estimators to become unreliable for establishing coarse correspondences. As shown in~\Cref{fig:fail}, in the fast-dancing sequence from the DNA-Rendering dataset~\cite{2023dnarendering}, we reduce the frame rate to $7.5$, effectively halving the original FPS. While our algorithm faithfully reconstructs frames $i$ and $i+1$, where ground-truth supervision is available, the interpolated intermediate frame at $i + 0.5$ exhibits noticeable artifacts due to the unreliable motion cues, which incorrectly associate part of the front leg in frame $i$ with the back leg in frame $i+1$. A similar issue appears in our Stage-Capture dataset, where reducing the frame rate to approximately $7.33$ leads to visible artifacts around the hand region in the intermediate frame at $i + 0.33$.
In practice, both FPS and physical motion speed contribute to inter-frame motion. Empirically, we find that for 1K videos, our method handles inter-frame motion up to around 50 pixels.
A promising direction for future work is to incorporate stronger motion priors or adopt alternative 4D representations that can robustly operate under such conditions, effectively treating the task as a 4D multi-frame start-end interpolation problem.

At discrete input frames, although we independently supervise adjacent sets of dynamic primitives with the ground truth to enforce consistency and smoothly transition their temporal opacity, their inherently disjoint nature can still cause slight flickering artifacts. Addressing this with a unified 4D representation remains future work.

\section{More Experiment Results}
\subsection{Per Scene Breakdown}
In~\Cref{tab:per_scene}, we provide the per-scene
breakdown of the quantitative results on the Stage-Capture Dataset.
\begin{table*}[t]
\caption{
\textbf{Per-scene quantitative comparisons on the Stage-Capture Dataset}
(foreground region). Higher PSNR/SSIM and lower LPIPS are better.
}
\vspace{-0.5em}
\centering
\setlength{\tabcolsep}{3pt} 
\begin{tabular}{lccccccccc c}
\toprule
& \multicolumn{9}{c}{Scenes} & \multirow{2}{*}{Avg.} \\
\cmidrule(lr){2-10}
Method & Bear & Doll & Undress & Open Case & Pass Doll &
Pickup Doll & Pack Computer & Stretch & Walk & \\
\midrule
\multicolumn{11}{c}{PSNR $\uparrow$} \\
\midrule
Deform-GS~\cite{wu20244d} & \tbest{29.91} & \tbest{27.21} & 29.46 & 29.99 & 27.87 & 28.30 & 29.96 & 25.81 & 27.56 & 28.45 \\
STGS~\cite{li2024spacetime} & 22.96 & 23.49 & 28.41 & 29.09 & 24.07 & 23.44 & 28.75 & 23.42 & 24.46 & 25.34 \\
GaussianFlow~\cite{gao2024gaussianflow} & 24.53 & 24.48 & 28.58 & 28.87 & 25.11 & 24.67 & 28.71 & 23.88 & 24.38 & 25.91 \\
2D Lifting~\cite{reda2022film,li2024spacetime} & 28.60 & 26.60 & \tbest{30.23} & \sbest{31.16} & \tbest{27.97} & \tbest{28.37} & \tbest{30.09} & \tbest{27.21} & \tbest{28.82} & \tbest{28.79} \\
\textbf{Ours} & \sbest{31.27} & \sbest{27.99} & \sbest{30.30} & \tbest{30.54} & \sbest{29.89} & \sbest{30.64} & \sbest{30.17} & \sbest{29.96} & \sbest{29.99} & \sbest{30.08} \\
\midrule
\multicolumn{11}{c}{SSIM $\uparrow$} \\
\midrule
Deform-GS~\cite{wu20244d} & 0.877 & 0.872 & 0.845 & 0.897 & 0.831 & 0.873 & 0.900 & 0.861 & 0.849 & 0.867 \\
STGS~\cite{li2024spacetime} & 0.817 & 0.820 & 0.831 & 0.890 & 0.780 & 0.818 & 0.883 & 0.811 & 0.778 & 0.825 \\
GaussianFlow~\cite{gao2024gaussianflow} & 0.824 & 0.826 & 0.833 & 0.887 & 0.789 & 0.827 & 0.884 & 0.806 & 0.753 & 0.825 \\
2D Lifting~\cite{reda2022film,li2024spacetime} & \tbest{0.880} & \tbest{0.875} & \tbest{0.875} & \sbest{0.916} & \tbest{0.859} & \tbest{0.891} & \tbest{0.907} & \tbest{0.888} & \tbest{0.878} & \tbest{0.886} \\
\textbf{Ours} & \sbest{0.905} & \sbest{0.899} & \sbest{0.878} & \tbest{0.914} & \sbest{0.877} & \sbest{0.912} & \sbest{0.912} & \sbest{0.925} & \sbest{0.911} & \sbest{0.904} \\
\midrule
\multicolumn{11}{c}{LPIPS $\downarrow$} \\
\midrule
Deform-GS~\cite{wu20244d} & \tbest{0.0329} & \tbest{0.0221} & 0.0347 & \tbest{0.0213} & \tbest{0.0481} & 0.0302 & \sbest{0.0148} & 0.0239 & \tbest{0.0171} & 0.0272 \\
STGS~\cite{li2024spacetime} & 0.0435 & 0.0303 & 0.0368 & 0.0235 & 0.0658 & 0.0449 & 0.0179 & 0.0337 & 0.0247 & 0.0357 \\
GaussianFlow~\cite{gao2024gaussianflow} & 0.0411 & 0.0287 & 0.0362 & 0.0231 & 0.0601 & 0.0411 & 0.0176 & 0.0309 & 0.0259 & 0.0339 \\
2D Lifting~\cite{reda2022film,li2024spacetime} & 0.0335 & 0.0229 & \sbest{0.0313} & \sbest{0.0190} & 0.0488 & \tbest{0.0298} & \tbest{0.0155} & \tbest{0.0228} & \tbest{0.0171} & \tbest{0.0267} \\
\textbf{Ours} & \sbest{0.0300} & \sbest{0.0197} & \tbest{0.0323} & \sbest{0.0190} & \sbest{0.0419} & \sbest{0.0210} & 0.0157 & \sbest{0.0118} & \sbest{0.0110} & \sbest{0.0225} \\
\bottomrule
\end{tabular}
\vspace{-1em}
\label{tab:per_scene}
\end{table*}

\subsection{Video-Based Slow-Motion Comparison}
\label{sec:supplementary:comparison}
The effectiveness of our method is clearly observed from the videos on the project page, where for each pair of input frames we interpolate 29 intermediate frames, slowing down the motion so that artifacts produced by other methods become more apparent and are otherwise difficult to capture in quantitative tables and qualitative static images.

In videos on the project page, all baseline methods, including 4D primitive–based approaches (with or without forward optical flow)~\cite{gao2024gaussianflow, li2024spacetime} and the 2D-to-3D lifted interpolation method~\cite{reda2022film}, produce noticeable ghosting artifacts for unseen temporal novel frames, especially in regions with large motion. For the 4D primitive–based methods~\cite{gao2024gaussianflow, li2024spacetime}, these artifacts appear in all frames except those with input ground-truth supervision. For the 2D interpolation–to–3D method~\cite{reda2022film}, ghosting occurs in all frames except the input-supervised frames and the directly interpolated middle frames. It is worth noting that in regions with small motion, these methods can indeed recover the correct primitive trajectories, as shown in several prior works. In some cases, because 2D interpolation effectively reduces the apparent inter-frame motion, the ghosting artifacts in unseen frames are also reduced, though not fully eliminated.

For the deformation-based method~\cite{wu20244d}, as discussed in~\Cref{sec:comparison}, using a single set of primitives to represent all frames in a dynamic scene enables it to capture a roughly correct global trajectory compared to STGS~\cite{li2024spacetime}. Nevertheless, in regions with fast motion, complex textures, or visibility changes, establishing detailed correspondences becomes challenging. As a result, we often observe parts of the scene being mapped to incorrect corresponding regions, leading to erroneous trajectories. Moreover, because the method relies solely on RGB cues to infer correspondences, resolving fine-grained textures becomes difficult, causing many detailed regions to appear blurry or distorted. This limitation can be observed in scenes where the intermediate frames appear visually correct overall but still achieve poor quantitative performance, as shown in~\Cref{tab:per_scene}, such as the Open Case scene.

\subsection{Video-Based Trajectory Comparison}
As discussed in~\Cref{sec:ablation}, using a spline trajectory produces smoother results than using a linear trajectory. In addition to the quantitative results shown in~\Cref{table:ablation}, we include a video comparison on the project page featuring a circular motion, which clearly reveals the piecewise-linear artifacts that appear when the spline design is omitted.

\subsection{Additional Analysis of Dynamic Stretching}
We discussed the benefits of introducing dynamic stretching in~\Cref{sec:stretch_and_relocate}: it prevents redundant representations of static objects across multiple primitive sets and allows more primitives to be allocated to dynamic regions. In addition, stretching the duration of static primitives offers another important advantage. Static regions that are covered by fewer camera views, and would therefore receive only sparse training signals, can instead accumulate supervision across multiple timesteps. This effectively increases their training signals, reduces flickering artifacts, and leads to more stable training.

In the main paper, \Cref{fig:ablation_stretching} provides a qualitative visualization of the automatically detected static and dynamic components. Here, we additionally present a quantitative analysis of this primitive reduction using an example scene (\Cref{fig:rebuttalflow}) from the DNA-Rendering Dataset~\cite{2023dnarendering}. Training under a 1M primitive budget with our modified MCMC strategy, we find that approximately 88K primitives (9\% of the total) are static Gaussians that span multiple frames. This brings the ``temporal sum'' of active Gaussians (i.e., the sum of all Gaussian temporal durations) to approximately 2.26M. Therefore, by applying dynamic stretching, our method reduces the effective number of primitives by a factor of $\mathbf{2.26\times}$.


\subsection{Video Results for Full Ablation Study}

On the project page, we additionally provide slow-motion playback videos for the remaining ablation experiments.

\begin{figure*}[t]
    \centering
    \includegraphics[width=\linewidth]{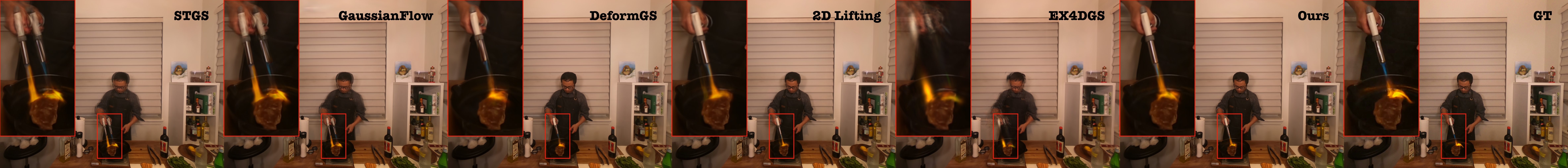}
    \vspace{-1.5em} 
    \caption{Qualitative comparison on Neural3DV: Scene with fire (rapid opacity changes) and non-stage-capture setting.}
    \label{fig:qualitative_neural3dv}
\end{figure*}

\subsection{Additional Comparison with Ex4DGS}
Since Ex4DGS~\cite{lee2024fully} also explicitly models temporal opacity and interpolates motion, we include it as another baseline among 4D primitive-based methods. Ex4DGS parameterizes temporal opacity as a constant window with Gaussian fall-offs at the boundaries while interpolating motion parameters across keyframes. Crucially, it leaves this temporal opacity unregularized. Consequently, the Gaussian fall-offs can become excessively narrow and close together, leading to severe overfitting when temporal signals are sparse. We quantitatively outperform Ex4DGS on the Stage-Capture dataset (\Cref{tab:stagecapture_split}) and qualitatively demonstrate that it suffers from ghosting artifacts similar to those in STGS~\cite{li2024spacetime} in the next subsection (\Cref{fig:qualitative_neural3dv}). Note that we evaluate Ex4DGS directly using its public codebase, unlike STGS and Deform-GS~\cite{wu20244d}, which benefit from integration into our framework with the MCMC strategy~\cite{kheradmand20243d}.

\begin{table}[t]
\centering
\caption{Quantitative results on the Stage-Capture dataset, including the Ex4DGS baseline.}

\begin{tabular}{l ccc}
\toprule
Method & PSNR$\uparrow$ & SSIM$\uparrow$ & LPIPS$\downarrow$ \\
\midrule
Deform-GS~\cite{wu20244d}    & 28.45 & 0.867 & 0.0272 \\
STGS~\cite{li2024spacetime}         & 25.34 & 0.825 & 0.0357 \\
GaussianFlow~\cite{gao2024gaussianflow} & 25.91 & 0.825 & 0.0339 \\
Ex4DGS~\cite{lee2024fully}       & 25.95 & 0.811 & 0.0379 \\
2D Lifting~\cite{reda2022film, li2024spacetime}   & \tbest{28.79} & \tbest{0.886} & \tbest{0.0267} \\
\midrule
Ours         & \sbest{30.08} & \sbest{0.904} & \sbest{0.0225} \\
\bottomrule
\end{tabular}
\label{tab:stagecapture_split}
\end{table}

\subsection{Results on Neural3DV Dataset}
Because our method regularizes temporal opacity, its behavior in scenes with rapidly changing opacity, such as fire, warrants discussion. Furthermore, to evaluate its generalization as a general reconstruction method, we test it on non-stage-captured data. Specifically, we compare our approach against the main paper baselines and Ex4DGS~\cite{lee2024fully} introduced in the previous section on the Flame Steak and Flame Salmon sequences from Neural3DV~\cite{li2022neural}, which feature both complex opacity changes and non-stage environments. Following our standard protocol, we subsample the videos to 1/10th of their original frame rate (from 30 FPS to 3 FPS), train using all cameras, and evaluate on all held-out temporal novel frames, per our setting. As shown in \Cref{fig:qualitative_neural3dv}, although our method is not explicitly designed for rapidly changing opacity, it performs reasonably well on fire scenes and generalizes effectively. \Cref{tab:neural3dv_split} provides the quantitative results; however, because these scenes are largely static, we believe the qualitative differences to be more revealing than the numerical metrics.

\begin{table}[t]
\centering
\caption{Quantitative evaluation on the Flame Steak and Flame Salmon scenes from the Neural3DV dataset.}
\begin{tabular}{l ccc}
\toprule
Method & PSNR$\uparrow$ & SSIM$\uparrow$ & LPIPS$\downarrow$ \\
\midrule
Deform-GS~\cite{wu20244d}     & 31.79 & 0.952 & 0.081 \\
STGS~\cite{li2024spacetime}         & 32.52 & \tbest{0.959} & \tbest{0.079} \\
GaussianFlow~\cite{gao2024gaussianflow} & 31.89 & 0.957 & 0.082 \\
Ex4DGS~\cite{lee2024fully}       & 31.06 & 0.919 & 0.094 \\
2D Lifting~\cite{reda2022film, li2024spacetime}   & \tbest{33.17} & \sbest{0.960} & 0.080 \\
\midrule
Ours         & \sbest{33.22} & \tbest{0.959} & \sbest{0.074} \\
\bottomrule
\end{tabular}
\label{tab:neural3dv_split}
\end{table}

\subsection{Discussion on Improved Optical Flow Quality}
Our method is robust to small errors in the pseudo-GT and sometimes even improves upon it, likely by leveraging multi-view information. As shown in \Cref{fig:rebuttalflow}, we observe improvements on fine-grained details, such as the hat tassel and the edges of the dress decorations, which are often blurry or inaccurate in the pseudo-ground truth estimations.

{\centering
\includegraphics[width=\linewidth]{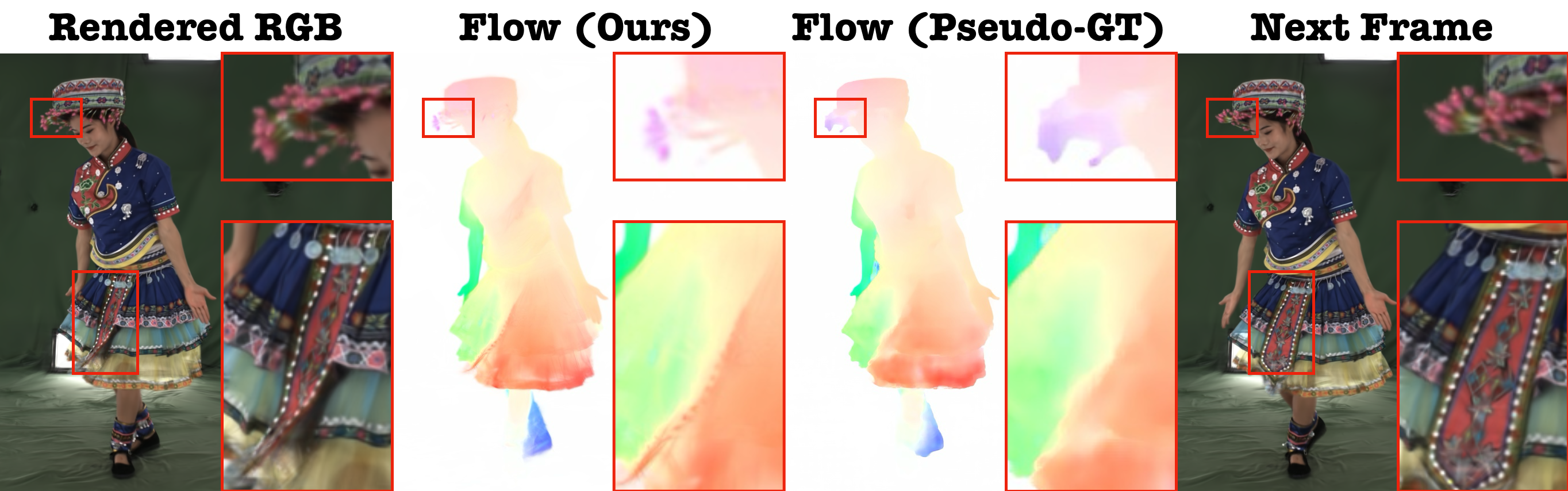}
\captionof{figure}{Comparison on rendered and pseudo-GT flow map.
}
\label{fig:rebuttalflow}
}

\subsection{Ablations on Other Optical Flow Methods}
Although our method exhibits robustness to small errors in the pseudo-ground-truth, the quality of the initial pseudo-GT remains critical, as it provides the essential rough correspondences. Without sufficiently accurate initialization, optimization tends to become trapped in poor local minima.
To investigate this, we conduct an ablation study by replacing our default flow estimator with the off-the-shelf SEA-RAFT~\cite{wang2024sea} for bidirectional flow supervision. Quantitative results are reported in \Cref{tab:flow_ablation}.

\begin{table}[t]
\centering
\caption{Ablation study on flow supervision: comparison of WAFT~\cite{wang2025waftwarpingalonefieldtransforms} and SEA-RAFT~\cite{wang2024sea} as optical flow supervision modules.}
\label{tab:flow_ablation}
\begin{tabular}{lccc}
\toprule
Method    & PSNR$\uparrow$ & SSIM$\uparrow$ & LPIPS$\downarrow$ \\
\midrule
WAFT~\cite{wang2025waftwarpingalonefieldtransforms}      & 30.08          & 0.904          & 0.0225 \\
SEA-RAFT~\cite{wang2024sea}  & 29.73          & 0.898          & 0.0253 \\
\bottomrule
\end{tabular}
\end{table}

\subsection{Memory and Training Time Footprint}
Although only a single rendering pass is required during inference, the triple rendering and flow supervision in our method increase training overhead. We report the average training efficiency and peak GPU memory usage on the DNA-Rendering Dataset~\cite{2023dnarendering}. Because our MCMC strategy maintains a fixed total primitive count, we compare our approach against STGS~\cite{li2024spacetime} under a shared budget of 1M primitives. We chose this baseline because both methods allow primitives to dynamically appear and disappear over time, unlike deformation-based methods where all primitives persist across all frames (which typically requires fewer total primitives).

\begin{table}[h]
\centering
\caption{Comparison of training efficiency and peak GPU memory under a shared budget of 1M primitives.}
\label{tab:efficiency}
\begin{tabular}{lcc}
\toprule
\textbf{Method} & \textbf{Training Time (s)} & \textbf{Peak Memory (GB)} \\
\midrule
STGS~\cite{li2024spacetime} & 1406.8 & 2.47 \\
Ours & 3794.3 & 3.14 \\
\bottomrule
\end{tabular}
\end{table}
